\documentclass{article}

\usepackage{microtype}
\usepackage{graphicx}
\usepackage{subfigure}
\usepackage{booktabs} 

\usepackage{hyperref}

\usepackage[accepted]{icml2023}

\usepackage{amsmath}
\usepackage{amssymb}
\usepackage{mathtools}
\usepackage{amsthm}

\usepackage[capitalize,noabbrev]{cleveref}

\usepackage[export]{adjustbox} 

\theoremstyle{plain}

\theoremstyle{definition}

\theoremstyle{remark}

\usepackage[textsize=tiny]{todonotes}

\begin{document}

\twocolumn[
\icmltitle{Addressing Discrepancies in Semantic and Visual Alignment in Neural Networks}




\begin{icmlauthorlist}
\icmlauthor{Natalie Abreu}{ll}
\icmlauthor{Nathan Vaska}{ll}
\icmlauthor{Victoria Helus}{ll}
\end{icmlauthorlist}

\icmlaffiliation{ll}{MIT Lincoln Laboratory, Lexington, MA, USA}

\icmlcorrespondingauthor{Nathan Vaska}{nathan.vaska@ll.mit.edu}

\icmlkeywords{Machine Learning, ICML}

\vskip 0.3in
]




\begin{abstract}
For the task of image classification, neural networks primarily rely on visual patterns. In robust networks, we would expect for visually similar classes to be represented similarly. We consider the problem of when semantically similar classes are visually dissimilar, and when visual similarity is present among non-similar classes. We propose a data augmentation technique with the goal of better aligning semantically similar classes with arbitrary (non-visual) semantic relationships. We leverage recent work in diffusion-based semantic mixing to generate semantic hybrids of two classes, and these hybrids are added to the training set as augmented data. We evaluate whether the method increases semantic alignment by evaluating model performance on adversarially perturbed data, with the idea that it should be easier for an adversary to switch one class to a similarly represented class. Results demonstrate that there is an increase in alignment of semantically similar classes when using our proposed data augmentation method.
\end{abstract}


\section{Introduction}
Within the common task of image classification, neural networks must rely on visual patterns in the images. While semantic relationships often follow from visual alignment, visuals and semantics are not always correlated. For instance, in a system that aims to distinguish child-safe objects from hazardous entities, a harmless object such as a spoon may appear visually similar to a dangerous object such as a knife, and a confusion between the two could have harmful implications. This example highlights the idea of mistake severity in neural networks -- most performance measures of classification models treat all errors equally, but in reality some errors are worse than others. Despite visual similarity, commonly confusing a knife for a spoon would be reason for high distrust in a system that is used to discriminate between harmful and safe objects. To address this concern, we propose a data augmentation method to incorporate prior semantic knowledge into the training process. In particular, we focus on the case of when semantic alignment is at odds with visual similarity, as this is where purely data-driven learning lacks crucial information of object semantics; visual information alone may fail to communicate class relationships.

With this method, we aim to increase alignment between semantically similar objects despite a lack of visual similarity. To measure this, we consider the metric of mistake severity over perturbed conditions, with the idea being that a model will be more likely to mistake one class for another if the classes are similarly represented.

The contributions of this work are as follows:
\begin{itemize}
    \item We propose a method of data augmentation using diffusion-based semantic mixing to increase alignment between semantically similar classes
    \item We construct a dataset with arbitrary class relationships based on CIFAR100 to evaluate our method when visual similarity is at odds with semantic similarity
    \item We evaluate our method on mistake severity over adversarially perturbed conditions and find that our data augmentation succeeds in increasing alignment between semantically similar classes
\end{itemize}

\section{Related Work}

Previous work has considered methods of incorporating semantic information into training, with methods such as introducing hierarchical loss functions (\cite{bertinetto}, \cite{zhao}, \cite{verma}, \cite{wu2016}) and aligning classes using adversarial perturbations (\cite{ma_har}, \cite{abreu2022addressing}). The notion of \textit{mistake severity} arises in many of these works as an alternate measure of model robustness, with the idea being that a mistake between classes that are highly dissimilar is worse than a mistake between semantically similar classes. \cite{bertinetto} notes that the improvement in the metric of mistake severity has been stagnant in recent years and argues that the metric should be revisited.  

Of particular interest in \cite{bertinetto} is a discussion in which the authors randomize the class relationships such that semantic proximity does not reflect visual similarity. In this setting, the performance of the hierarchical methods considered deteriorates, suggesting that the visual similarity of related classes in one's hierarchy is essential to the success of the proposed methods. The authors note, ``while one may wish to enforce application-specific relationships using this approach…, the effectiveness of doing so may be constrained by underlying properties of the data" (\cite{bertinetto}). The work in \cite{abreu2022addressing} finds similar behavior when visual relationships no longer support semantic ones. We aim to address this dependency on visual similarity in our data augmentation method.

Additionally, there has been prior work in using diffusion models to generate synthetic training data. \cite{syndata} uses diffusion models to provide synthetic data for image classification. \cite{he2023} explores the use of synthetic data generated from the text-to-image generation model GLIDE (\cite{glide}) in zero-shot and few-shot settings, as well as for model pre-training. They find that synthetic data can be beneficial in these settings, and further investigate strategies to increase data diversity and reduce data noise for synthetic data generation. Similar to our approach, \cite{augdiffusion} proposes a diffusion-based data augmentation method.  \cite{augdiffusion} uses diffusion models to augment individual images to diversify high-level semantic attributes of images; for instance, modifying the appearance of the face of a truck or the landscape of the background. Our work differs in that we apply our augmentation to create semantic hybrids of images rather than to diversify samples of a given class.



In our method, we utilize semantic perturbations of the training samples as a way of incorporating semantic knowledge. Specifically, we use semantic mixing of training samples, a recent task that aims to blend two different concepts to synthesize a new concept. \cite{liew2022magicmix} present a method called MagicMix to semantically mix concepts based on pre-trained text-conditioned diffusion models. MagicMix does not require any spatial mask or re-training, which made it lightweight enough to use in our method. 

We use adversarial perturbations in our evaluation to provide insight on how the model aligns the representation of classes. Adversarial perturbations, as introduced in \cite{szegedy}, are small perturbations that can change the model's prediction of an image. \cite{madry} provides an optimization view of adversarial perturbations that allows us to solve for attacks of an $l_2$-bounded projected gradient descent (PGD) adversary.

\section{Method}

\begin{figure}
    \centering
    \includegraphics[width=80mm]{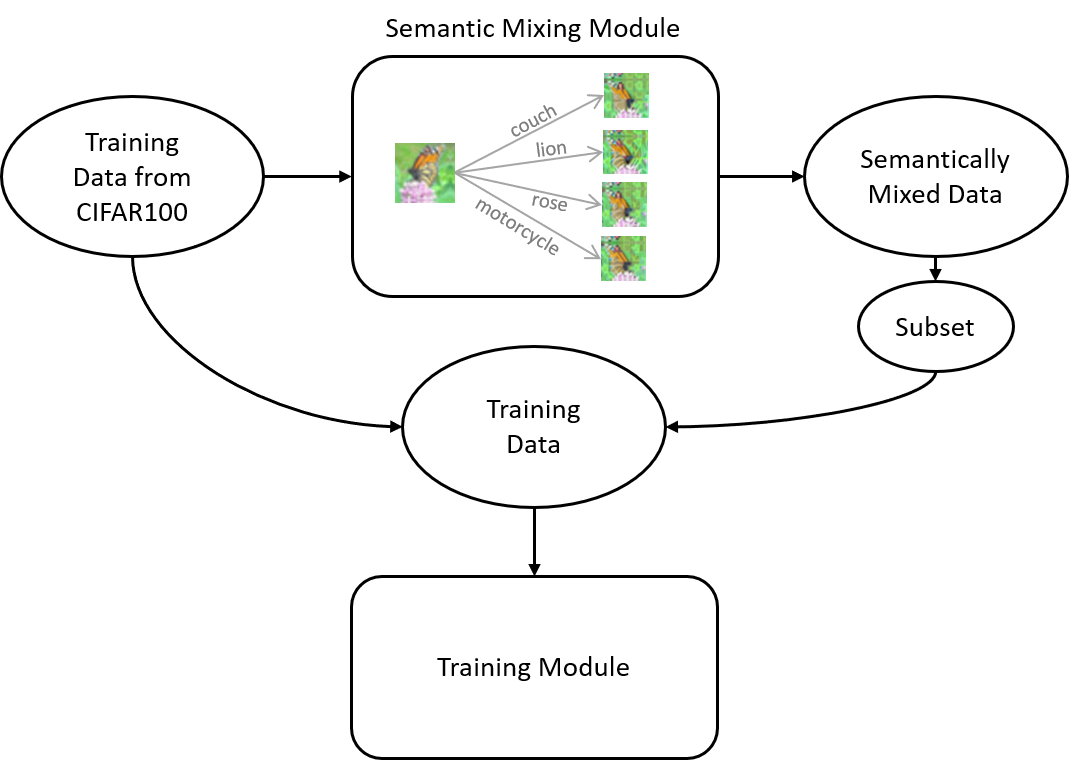}
    \caption{Diagram of method. The semantic mixing module is illustrated with an example used in our experiment, where ``Butterfly", ``Couch", ``Lion", ``Rose", and ``Motorcycle" are grouped as an arbitrary super class. In the semantic mixing module, an instance of a butterfly is used to generate new hybrid images. A subset of these mixed images is added to the original clean training data from CIFAR100 to form the final augmented set of training data. }
    \label{fig:diagram}
\end{figure}

We embed semantic knowledge into the training process by incorporating ``semantically mixed" data in the training process. Specifically, we propose a data augmentation technique in which the training data is used to generate new training samples which are hybrids of two semantically similar classes. For efficiency, we pre-generate this data using the MagicMix pipeline: for each image in the training set, we generate a new mixed image towards each other class in its super class (See Figure \ref{fig:diagram}).

\begin{figure}
    \centering
    \includegraphics[width=80mm]{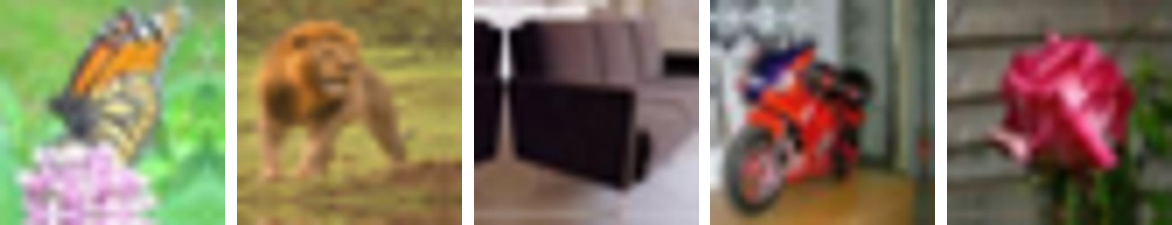}
    \caption{Five images in super class C, illustrating visual dissimilarity between images in the same ``semantic" super class.}
    \label{fig:example_classes}
\end{figure}

\begin{table*}[t]
\centering
\begin{tabular}{ |c|c c c c c| }
 \hline
  & Flowers & Furniture & Insects & Carnivores & Vehicles  \\ 
 \hline
 A & Orchid & Bed & Bee & Bear & Bicycle \\ 
 B & Poppy & Chair & Beetle & Leopard & Bus \\ 
 C & Rose & Couch & Butterfly & Lion & Motorcycle \\
 D & Sunflower & Table & Caterpillar & Tiger & Pickup truck \\
 E & Tulip & Wardrobe & Cockroach & Wolf & Train \\
 \hline
\end{tabular}
\caption{Chart depicting refactored superclasses. The columns depict the original super classes ({Flowers, Furniture, Insects, Carnivores, Vehicles}), and the rows depict the new super classes ({A,B,C,D,E}). Note that there is high visual similarity within classes in the same column, but not in the same row.}
\label{table1}
\end{table*}

We vary the amount of augmented data used in training by specifying a probability of adding an augmented image of the class of any given instance encountered in training. Given the high ratio of augmented data to clean training data, this method allows us to prevent the augmented data from completely dominating the clean data. In our experiments, we refer to "low augmentation" as having a 25\% probability of adding an augmented image for any given instance in training, and "high augmentation" as having a 50\% probability of adding an augmented image. The augmented image is chosen by randomly selecting a pre-generated image with the same base class as the given instance. Augmented instances are labelled 50\% as the base class and 50\% as the target class.

\section{Experiments}
With the focus of adding alignment within arbitrary class relationships, we formed our dataset to minimize visual similarity of classes within the same super class and incorporate visual similarity of classes in different super classes. As our dataset, we selected five visually dissimilar super classes from CIFAR100 (\cite{cifar100}) and redistributed classes such that one class from each original super class was in each of the new super classes. Our super class groupings are shown in Table \ref{table1}. We will refer to the original super classes ({Flowers, Furniture, Insects, Carnivores, Vehicles}) as \textit{visual} super classes and the new super classes (A, B, C, D, E) as \textit{semantic} super classes to avoid confusion between the two.

\begin{table}
\centering
\setlength{\tabcolsep}{1pt}
\begin{tabular}{lllllll}
\includegraphics[width=.16\linewidth,valign=m]{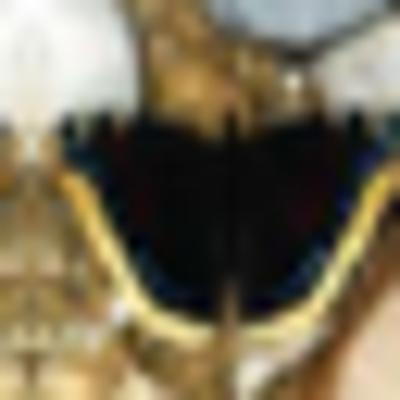} & \includegraphics[width=.16\linewidth,valign=m]{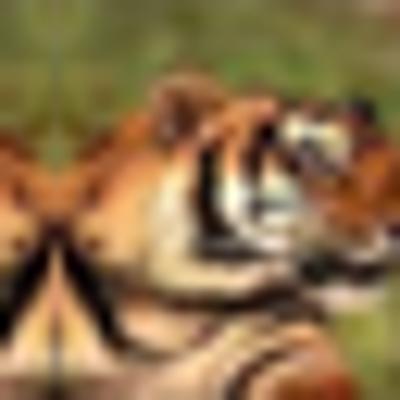} &
\includegraphics[width=.16\linewidth,valign=m]{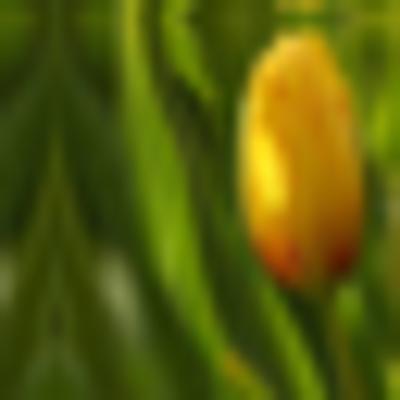} &
\includegraphics[width=.16\linewidth,valign=m]{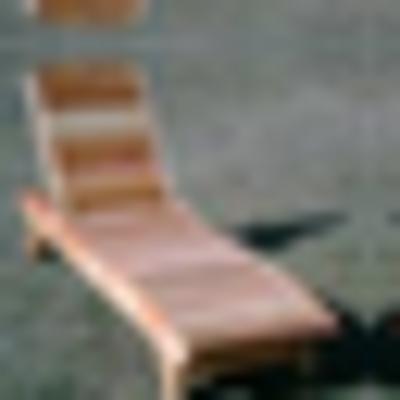} &
\includegraphics[width=.16\linewidth,valign=m]{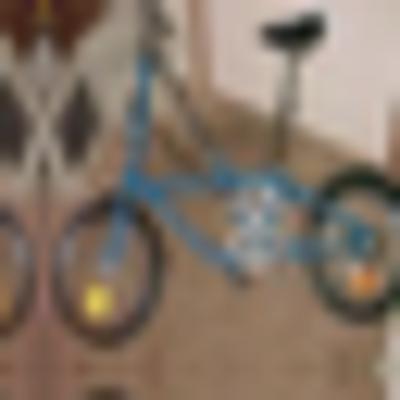} \\
\includegraphics[width=.16\linewidth,valign=m]{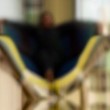} & \includegraphics[width=.16\linewidth,valign=m]{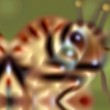} &
\includegraphics[width=.16\linewidth,valign=m]{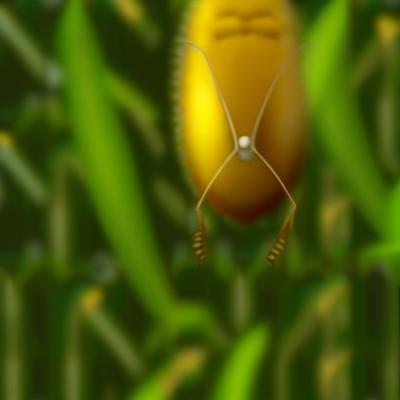} &
\includegraphics[width=.16\linewidth,valign=m]{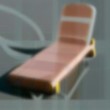} &
\includegraphics[width=.16\linewidth,valign=m]{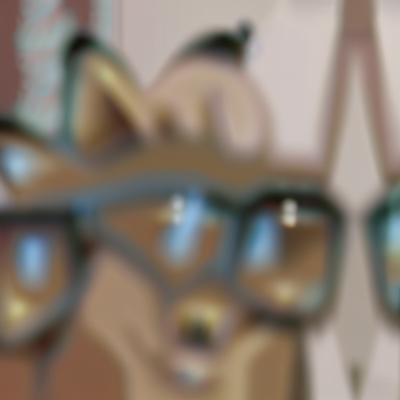} \\
\includegraphics[width=.16\linewidth,valign=m]{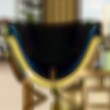} & \includegraphics[width=.16\linewidth,valign=m]{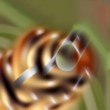} & \includegraphics[width=.16\linewidth,valign=m]{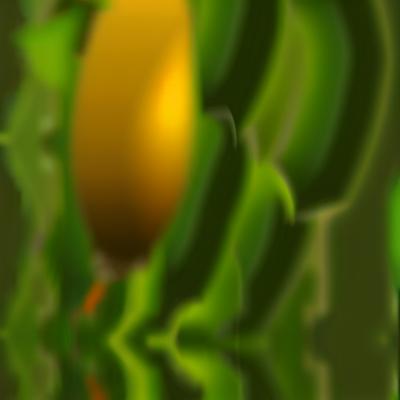} &
\includegraphics[width=.16\linewidth,valign=m]{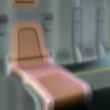} &
\includegraphics[width=.16\linewidth,valign=m]{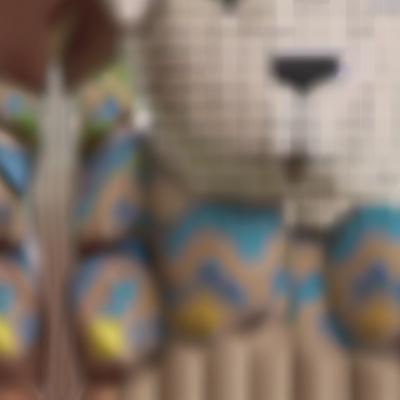} \\

\end{tabular}
\caption{Examples of generated images. From left to right: butterfly to couch, tiger to caterpillar, tulip to cockroach, chair to bus, bicycle to bear. The top row, middle row, and bottom row show the original image and images generated with a mix factor of 0.50 and 0.75 respectively.}
\label{table:hybridimages}
\end{table}

We pre-generate our hybrid images using the MagicMix pipeline from \cite{liew2022magicmix} using image-text mixing. For each image in the training set, we create four hybrid images, one hybrid image per each other fine class within the same super class. The image from the training set is used as the base image and the prompt in the MagicMix module is set to the fine class name in the same super class. The MagicMix module allows a mix factor in the range $[0,1]$ to be set to define the strength of the mixing towards the target prompt. We vary the mix factor over models - for low mix strength, we use a mix factor of 0.50 and for high mix strength, we use a mix factor of 0.75. Example images are shown in Table \ref{table:hybridimages}.

All models used a ResNet50 architecture as described in \cite{resnet} and were trained on the above dataset. We used a learning rate of 0.1, a batch size of 100, and standard values for remaining training parameters. Additional data augmentation was applied in the form of random cropping and random horizontal flipping. Models were trained for 100 epochs.

We evaluate our method using the metric of super class accuracy on mistakes. Particularly, we look at the metric over adversarial instances of increasing severity, with the idea being that, if the model has a similar representation of visually similar classes, it will be easy to find an adversarial perturbation from one class to another class outside of its super class.

Adversarial attacks are modeled with an $l_2$-bounded projected gradient descent adversary as proposed in \cite{madry}. For a model $f$ with learned parameters $\theta$ over a data distribution $D$ and loss function $\mathcal{L}$, we find an  adversarial perturbation $\delta$ of a given instance $x$ with label $y$ by solving
\[ max_{\delta: ||\delta||<\epsilon}\, \mathbb{E}_{(x,y)\sim D}[\mathcal{L}(f_\theta(x+\delta),y)].\]
where $\epsilon$ is the $l_2$ bound of the adversary.

The models we compared were as follows:
\begin{itemize}
    \item \textbf{Standard model}: Model trained with no additional augmented data
    \item \textbf{Low augmentation, low mix strength}: Model trained with 25\% additional augmented data and mix strength of 0.5
    \item 	\textbf{Low augmentation, high mix strength}: Model trained with 25\% additional augmented data and mix strength of 0.75
    \item \textbf{High augmentation, low mix strength}: Model trained with 50\% additional augmented data and mix strength of 0.5
    \item \textbf{High augmentation, high mix strength}: Model trained with 50\% additional augmented data and mix strength of 0.75
\end{itemize}

\section{Results}
In this section, we will show results on mistake severity over adversarial perturbations of increasing severity. First, we will demonstrate that the models with our proposed augmentation technique perform better in terms of mistake severity on adversarially perturbed instances. We will additionally demonstrate that our technique decreases mistakes across visually similar classes. These results indicate that our method helps to increase semantic alignment at odds with visual similarities.

\begin{figure}
    \centering
    \includegraphics[width=60mm]{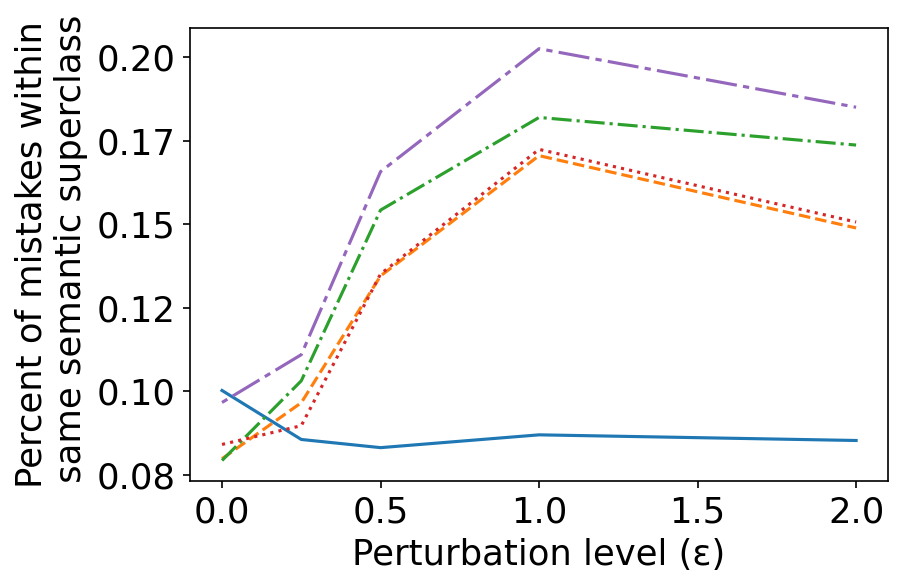}
    \includegraphics[width=60mm]{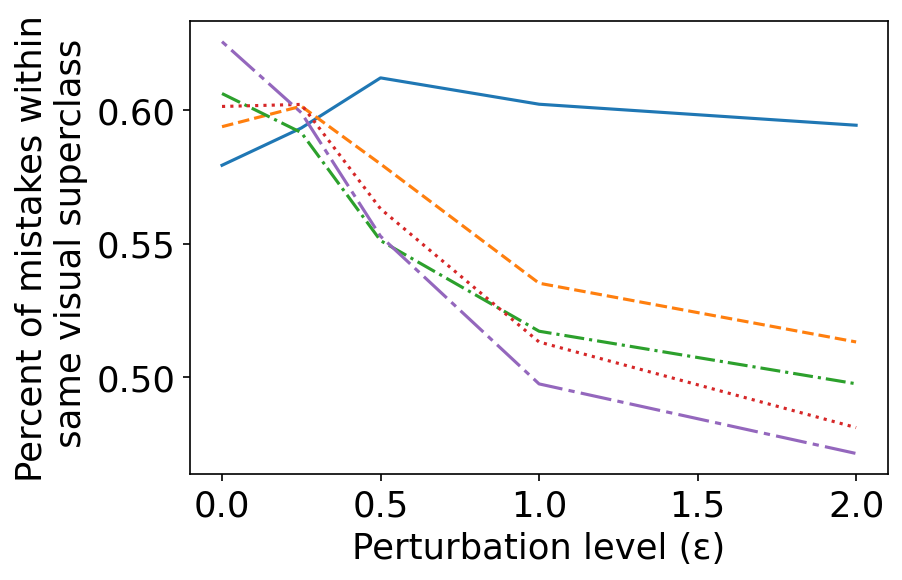}
    \includegraphics[width=60mm]{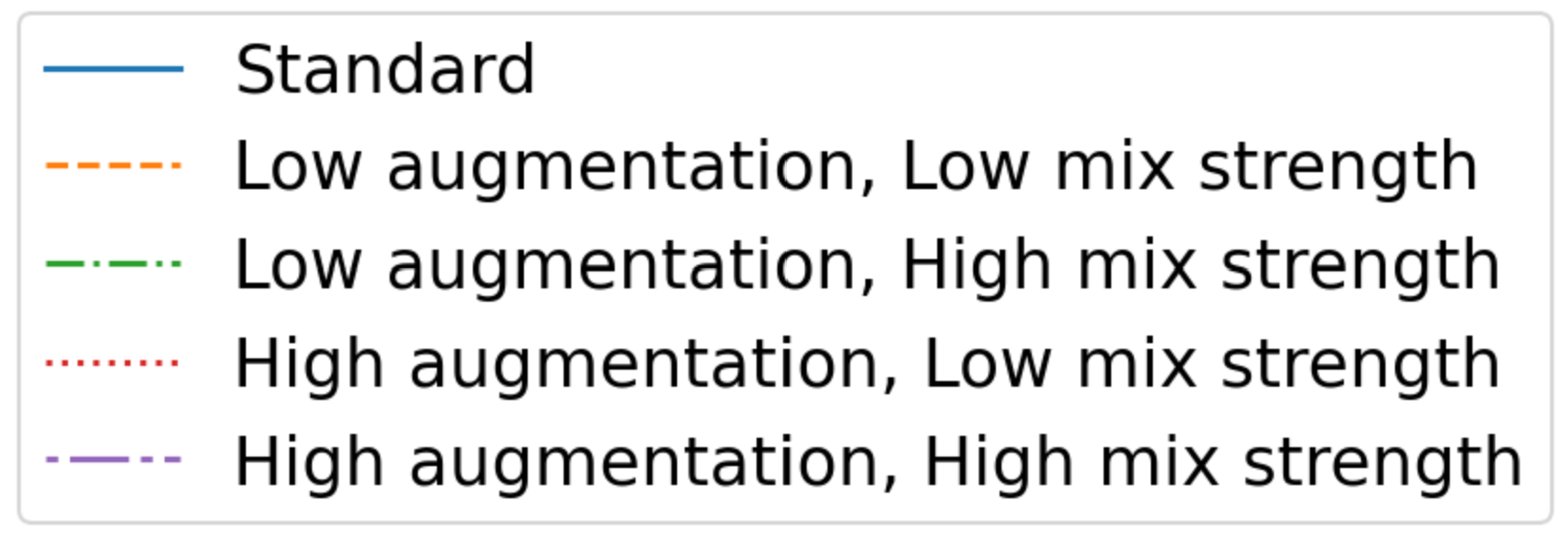}
    \caption{Top: Percent of mistakes that were made within the same semantic super class. For instance, a tulip getting mistaken for a wardrobe. Bottom: Percent of mistakes that were made within the same visual super class. For instance, a tulip getting mistaken for a rose.}
    \label{fig:combined_mistakes}
\end{figure}



The models using our data augmentation technique have considerably higher super class accuracy on mistakes on perturbed instances than the standard model, as seen in Figure \ref{fig:combined_mistakes}. The high data augmentation, high mix strength model performs best overall on this metric, with  performance close to the standard model on the clean data and best performance on all nonzero levels of perturbation. To address the similar performance of the standard and data augmentation models on the clean data, we posit that the simplicity of the CIFAR100 dataset causes the models to only makes mistakes on difficult examples (e.g., ones with unique or misleading features) at low levels of perturbation. As the perturbation level increases, the models may start to makes mistakes on examples with more standard features, which offers an explanation to the fact that better performance of the models with data augmentation is only present on more highly perturbed data.


The models using the data augmentation technique additionally have lower percents of mistakes between classes  in the same visual super classes (e.g. "Flowers") (shown in Figure \ref{fig:combined_mistakes}). This demonstrates that the model learn lower correlations between visually similar classes that are not given as semantically similar.

\begin{figure}
    \centering
    \includegraphics[width=60mm]{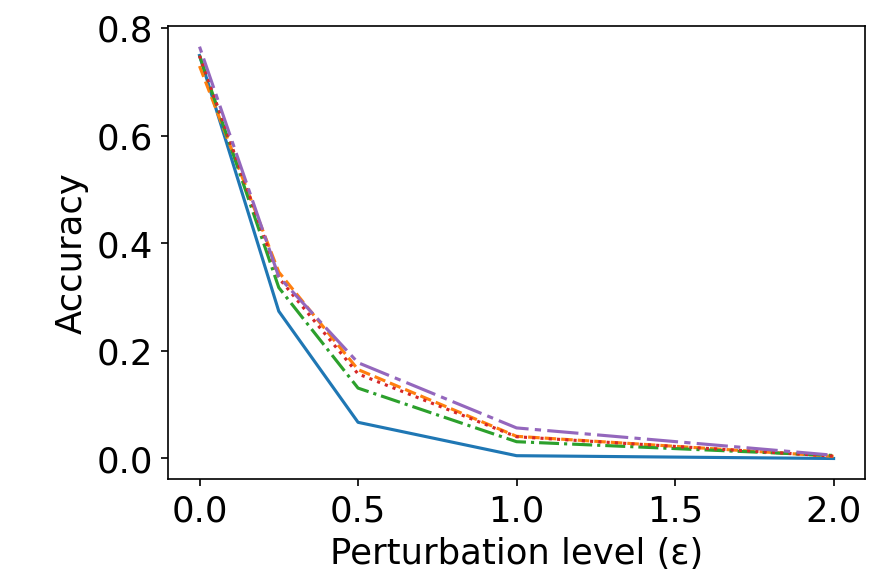}
    \includegraphics[width=60mm]{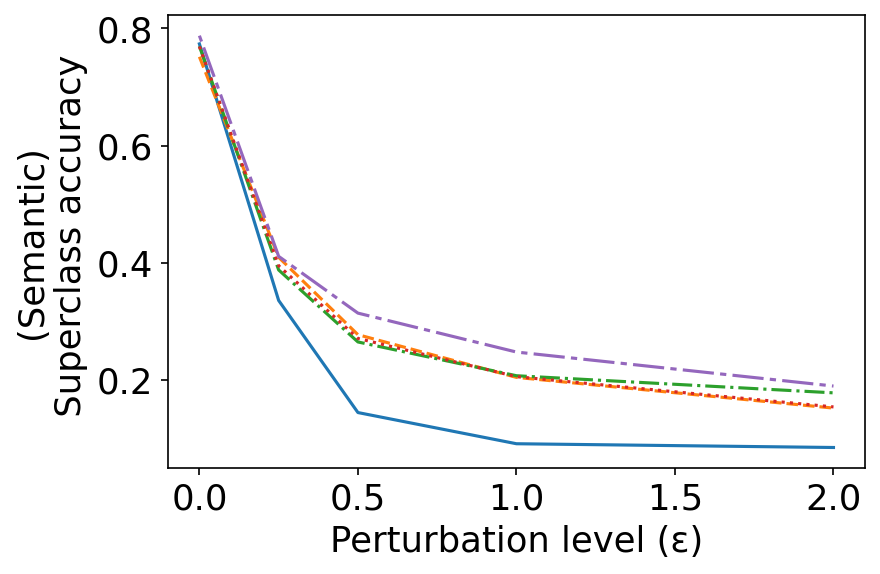}
    \includegraphics[width=60mm]{charts/legend.png}
    \caption{Top: Fine class accuracy over all test instances. Bottom: Super class accuracy overall (percent of test instances classified as a fine class in the correct semantic super class)}
    \label{fig:combined_accuracy}
\end{figure}

Finally, we show overall model accuracy and semantic super class accuracy in Figure \ref{fig:combined_accuracy}. The models using the data augmentation technique improve on both metrics over all non-zero levels of adversarial perturbation, and the high augmentation, high mix strength model additionally improves on clean accuracy and super class accuracy. As the dataset is not very challenging, the improvement of even the best performing model with data augmentation is marginal on clean accuracy and super class accuracy. As the data set gets more challenging with added perturbation, our method improves on performance as MagicMix distortions help group features of classes in the same semantic super class. Even at the highest level of perturbation, some semantic alignment is maintained in the models with data augmentation.




\section{Discussion and Conclusions}
Our findings give promising first results for using data augmentation as a method of increasing semantic alignment between classes with arbitrary visual relationships. More generally, this finding suggests potential for synthetic data to inject prior knowledge into training. As future work, we would like to apply our method to a more complex dataset, where the model is more likely to see ambiguous images or images otherwise more difficult to classify. Additionally, the method could be extended to an application with domain-specific knowledge that needs to be incorporated rather than arbitrary class relationships.

\section*{Acknowledgements}
DISTRIBUTION STATEMENT A. Approved for public release. Distribution is unlimited. This material is based upon work supported by the Department of the Air Force under Air Force Contract No. FA8702-15-D-0001. Any opinions, findings, conclusions or recommendations expressed in this material are those of the author(s) and do not necessarily reflect the views of the Department of the Air Force. 

© 2023 Massachusetts Institute of Technology. 

Delivered to the U.S. Government with Unlimited Rights, as defined in DFARS Part 252.227-7013 or 7014 (Feb 2014). Notwithstanding any copyright notice, U.S. Government rights in this work are defined by DFARS 252.227-7013 or DFARS 252.227-7014 as detailed above. Use of this work other than as specifically authorized by the U.S. Government may violate any copyrights that exist in this work.

The authors would like to thank Dr. Sanjeev Mohindra and Ms. Paula Donovan for their support.


\bibliography{main}

\begin{thebibliography}{15}
\providecommand{\natexlab}[1]{#1}
\providecommand{\url}[1]{\texttt{#1}}
\expandafter\ifx\csname urlstyle\endcsname\relax
  \providecommand{\doi}[1]{doi: #1}\else
  \providecommand{\doi}{doi: \begingroup \urlstyle{rm}\Url}\fi

\bibitem[Abreu et~al.(2022)Abreu, Vaska, and Helus]{abreu2022addressing}
Abreu, N., Vaska, N., and Helus, V.
\newblock Addressing mistake severity in neural networks with semantic
  knowledge.
\newblock \emph{CoRR}, abs/2211.11880, 2022.
\newblock \doi{10.48550/arXiv.2211.11880}.
\newblock URL \url{https://doi.org/10.48550/arXiv.2211.11880}.

\bibitem[Azizi et~al.(2023)Azizi, Kornblith, Saharia, Norouzi, and
  Fleet]{syndata}
Azizi, S., Kornblith, S., Saharia, C., Norouzi, M., and Fleet, D.~J.
\newblock Synthetic data from diffusion models improves imagenet
  classification.
\newblock \emph{CoRR}, abs/2304.08466, 2023.
\newblock \doi{10.48550/arXiv.2304.08466}.
\newblock URL \url{https://doi.org/10.48550/arXiv.2304.08466}.

\bibitem[Bertinetto et~al.(2020)Bertinetto, M{\"{u}}ller, Tertikas, Samangooei,
  and Lord]{bertinetto}
Bertinetto, L., M{\"{u}}ller, R., Tertikas, K., Samangooei, S., and Lord, N.~A.
\newblock Making better mistakes: Leveraging class hierarchies with deep
  networks.
\newblock In \emph{2020 {IEEE/CVF} Conference on Computer Vision and Pattern
  Recognition, {CVPR} 2020, Seattle, WA, USA, June 13-19, 2020}, pp.\
  12503--12512. Computer Vision Foundation / {IEEE}, 2020.
\newblock \doi{10.1109/CVPR42600.2020.01252}.
\newblock URL
  \url{https://openaccess.thecvf.com/content\_CVPR\_2020/html/Bertinetto\_Making\_Better\_Mistakes\_Leveraging\_Class\_Hierarchies\_With\_Deep\_Networks\_CVPR\_2020\_paper.html}.

\bibitem[He et~al.(2016)He, Zhang, Ren, and Sun]{resnet}
He, K., Zhang, X., Ren, S., and Sun, J.
\newblock Deep residual learning for image recognition.
\newblock In \emph{2016 {IEEE} Conference on Computer Vision and Pattern
  Recognition, {CVPR} 2016, Las Vegas, NV, USA, June 27-30, 2016}, pp.\
  770--778. {IEEE} Computer Society, 2016.
\newblock \doi{10.1109/CVPR.2016.90}.
\newblock URL \url{https://doi.org/10.1109/CVPR.2016.90}.

\bibitem[He et~al.(2022)He, Sun, Yu, Xue, Zhang, Torr, Bai, and Qi]{he2023}
He, R., Sun, S., Yu, X., Xue, C., Zhang, W., Torr, P. H.~S., Bai, S., and Qi,
  X.
\newblock Is synthetic data from generative models ready for image recognition?
\newblock \emph{CoRR}, abs/2210.07574, 2022.
\newblock \doi{10.48550/arXiv.2210.07574}.
\newblock URL \url{https://doi.org/10.48550/arXiv.2210.07574}.

\bibitem[Krizhevsky(2009)]{cifar100}
Krizhevsky, A.
\newblock Learning multiple layers of features from tiny images.
\newblock pp.\  32--33, 2009.
\newblock URL
  \url{https://www.cs.toronto.edu/~kriz/learning-features-2009-TR.pdf}.

\bibitem[Liew et~al.(2022)Liew, Yan, Zhou, and Feng]{liew2022magicmix}
Liew, J.~H., Yan, H., Zhou, D., and Feng, J.
\newblock Magicmix: Semantic mixing with diffusion models.
\newblock \emph{CoRR}, abs/2210.16056, 2022.
\newblock \doi{10.48550/arXiv.2210.16056}.
\newblock URL \url{https://doi.org/10.48550/arXiv.2210.16056}.

\bibitem[Ma et~al.(2021)Ma, Virmaux, Scaman, and Lu]{ma_har}
Ma, A., Virmaux, A., Scaman, K., and Lu, J.
\newblock Improving hierarchical adversarial robustness of deep neural
  networks.
\newblock \emph{CoRR}, abs/2102.09012, 2021.
\newblock URL \url{https://arxiv.org/abs/2102.09012}.

\bibitem[Madry et~al.(2018)Madry, Makelov, Schmidt, Tsipras, and Vladu]{madry}
Madry, A., Makelov, A., Schmidt, L., Tsipras, D., and Vladu, A.
\newblock Towards deep learning models resistant to adversarial attacks.
\newblock In \emph{6th International Conference on Learning Representations,
  {ICLR} 2018, Vancouver, BC, Canada, April 30 - May 3, 2018, Conference Track
  Proceedings}. OpenReview.net, 2018.
\newblock URL \url{https://openreview.net/forum?id=rJzIBfZAb}.

\bibitem[Nichol et~al.(2022)Nichol, Dhariwal, Ramesh, Shyam, Mishkin, McGrew,
  Sutskever, and Chen]{glide}
Nichol, A.~Q., Dhariwal, P., Ramesh, A., Shyam, P., Mishkin, P., McGrew, B.,
  Sutskever, I., and Chen, M.
\newblock {GLIDE:} towards photorealistic image generation and editing with
  text-guided diffusion models.
\newblock In Chaudhuri, K., Jegelka, S., Song, L., Szepesv{\'{a}}ri, C., Niu,
  G., and Sabato, S. (eds.), \emph{International Conference on Machine
  Learning, {ICML} 2022, 17-23 July 2022, Baltimore, Maryland, {USA}}, volume
  162 of \emph{Proceedings of Machine Learning Research}, pp.\  16784--16804.
  {PMLR}, 2022.
\newblock URL \url{https://proceedings.mlr.press/v162/nichol22a.html}.

\bibitem[Szegedy et~al.(2014)Szegedy, Zaremba, Sutskever, Bruna, Erhan,
  Goodfellow, and Fergus]{szegedy}
Szegedy, C., Zaremba, W., Sutskever, I., Bruna, J., Erhan, D., Goodfellow,
  I.~J., and Fergus, R.
\newblock Intriguing properties of neural networks.
\newblock In Bengio, Y. and LeCun, Y. (eds.), \emph{2nd International
  Conference on Learning Representations, {ICLR} 2014, Banff, AB, Canada, April
  14-16, 2014, Conference Track Proceedings}, 2014.
\newblock URL \url{http://arxiv.org/abs/1312.6199}.

\bibitem[Trabucco et~al.(2023)Trabucco, Doherty, Gurinas, and
  Salakhutdinov]{augdiffusion}
Trabucco, B., Doherty, K., Gurinas, M., and Salakhutdinov, R.
\newblock Effective data augmentation with diffusion models.
\newblock \emph{CoRR}, abs/2302.07944, 2023.
\newblock \doi{10.48550/arXiv.2302.07944}.
\newblock URL \url{https://doi.org/10.48550/arXiv.2302.07944}.

\bibitem[Verma et~al.(2012)Verma, Mahajan, Sellamanickam, and Nair]{verma}
Verma, N., Mahajan, D., Sellamanickam, S., and Nair, V.
\newblock Learning hierarchical similarity metrics.
\newblock In \emph{2012 {IEEE} Conference on Computer Vision and Pattern
  Recognition, Providence, RI, USA, June 16-21, 2012}, pp.\  2280--2287. {IEEE}
  Computer Society, 2012.
\newblock \doi{10.1109/CVPR.2012.6247938}.
\newblock URL \url{https://doi.org/10.1109/CVPR.2012.6247938}.

\bibitem[Wu et~al.(2016)Wu, Merler, Uceda{-}Sosa, and Smith]{wu2016}
Wu, H., Merler, M., Uceda{-}Sosa, R., and Smith, J.~R.
\newblock Learning to make better mistakes: Semantics-aware visual food
  recognition.
\newblock In Hanjalic, A., Snoek, C., Worring, M., Bulterman, D. C.~A., Huet,
  B., Kelliher, A., Kompatsiaris, Y., and Li, J. (eds.), \emph{Proceedings of
  the 2016 {ACM} Conference on Multimedia Conference, {MM} 2016, Amsterdam, The
  Netherlands, October 15-19, 2016}, pp.\  172--176. {ACM}, 2016.
\newblock \doi{10.1145/2964284.2967205}.
\newblock URL \url{https://doi.org/10.1145/2964284.2967205}.

\bibitem[Zhao et~al.(2011)Zhao, Fei{-}Fei, and Xing]{zhao}
Zhao, B., Fei{-}Fei, L., and Xing, E.~P.
\newblock Large-scale category structure aware image categorization.
\newblock In Shawe{-}Taylor, J., Zemel, R.~S., Bartlett, P.~L., Pereira, F.
  C.~N., and Weinberger, K.~Q. (eds.), \emph{Advances in Neural Information
  Processing Systems 24: 25th Annual Conference on Neural Information
  Processing Systems 2011. Proceedings of a meeting held 12-14 December 2011,
  Granada, Spain}, pp.\  1251--1259, 2011.
\newblock URL
  \url{https://proceedings.neurips.cc/paper/2011/hash/d5cfead94f5350c12c322b5b664544c1-Abstract.html}.

\end{thebibliography}
\bibliographystyle{icml2023}



\end{document}